\documentclass{bmvc2k}

\title{Towards Robust Few-shot Point Cloud Semantic Segmentation}

\addauthor{Yating Xu}{xu.yating@u.nus.edu}{1}
\addauthor{Na Zhao}{na_zhao@sutd.edu.sg}{2}
\addauthor{Gim Hee Lee}{gimhee.lee@nus.edu.sg}{1}

\addinstitution{
 Department of Computer Science \\
 National University of Singapore\\
 Singapore
}
\addinstitution{
 Singapore University of Technology and Design\\
 Singapore
}

\runninghead{Xu \etal}{Towards Robust Few-shot Point Cloud Semantic Segmentation}

\def\eg{\emph{e.g}\bmvaOneDot}

\def\etal{\emph{et al}\bmvaOneDot}
\newcommand{\ie}{\textit{i.e.}~}
\newcommand{\cf}{\textit{cf.}~}

\usepackage{amsmath,amssymb} 
\usepackage{color}
\usepackage{booktabs}
\usepackage{multirow}
\usepackage{tabularx,verbatim}
\usepackage{xspace}
\usepackage{algorithm}
\usepackage{algorithmic}
\usepackage{setspace}
\usepackage{comment}
\usepackage{bbm}
\usepackage{enumitem}
\usepackage{colortbl}
\usepackage{color, xcolor} 
\usepackage[T1]{fontenc}
\usepackage{bbm}
\usepackage{wrapfig}
\usepackage{bbding}

\newcommand{\ytComment}[1]{\textcolor{black}{#1}}
\begin{document}

\maketitle

\begin{abstract}
Few-shot point cloud semantic segmentation aims to train a model to quickly adapt to new unseen classes with only a handful of support set samples. However, the noise-free assumption in the support set can be easily violated in many practical real-world settings. In this paper, we focus on improving the robustness of few-shot point cloud segmentation under the detrimental influence of noisy support sets during testing time. To this end, we first propose a Component-level Clean Noise Separation (CCNS) representation learning to learn discriminative feature representations that separates the clean samples of the target classes from the noisy samples. Leveraging the well-separated clean and noisy support samples from our CCNS, we further propose a Multi-scale Degree-based Noise Suppression (MDNS) scheme to remove the noisy shots from the support set.
We conduct extensive experiments on various noise settings on two benchmark datasets. Our results show that the combination of CCNS and MDNS significantly improves the performance. 
Our code is available at \url{https://github.com/Pixie8888/R3DFSSeg}.

\end{abstract}

\section{Introduction}
\label{sec:intro}
Few-shot point cloud semantic segmentation (3DFSSeg) \cite{zhao2021few,mao2022bidirectional} is a pragmatic direction as it is able to segment novel classes during testing stage with only few labeled samples. In contrast to the fully-supervised methods \cite{wang2019dynamic,qi2017pointnet,qi2017pointnet++} which only work for close set, 3DFSSeg has better generalization ability. However, it assumes that the learning samples of the novel classes are correctly labeled during online testing time. 


Unfortunately, the assumption of completely clean data could be violated in practice due to a variety of reasons.
First, human labeling is error-prone. The irregular data structure, low-resolution, and subtle inter-class geometric difference make human annotators themselves hard to correctly recognize objects \cite{wang2019latte}. The crowdsourcing labeling further stresses the annotation quality \cite{walter2020evaluation}. As a consequence, ScanNet \cite{dai2017scannet} still contains annotation mistakes \cite{ye2021learning} after manual refinement over an extended period of time. 
Second, the industry is actively seeking cheaper and more efficient annotation system to replace human labeling, \eg semi-automatic labeling \cite{li20203d,wirth2019pointatme} and fully automatic annotation \cite{humblot2023cad,bloembergen2021automatic,chen2019image}. 
It further challenges the curation of high-quality data.

\begin{wrapfigure}{R}{0.51\textwidth}
\vspace{-4mm}
\centering
\includegraphics[scale=0.38]{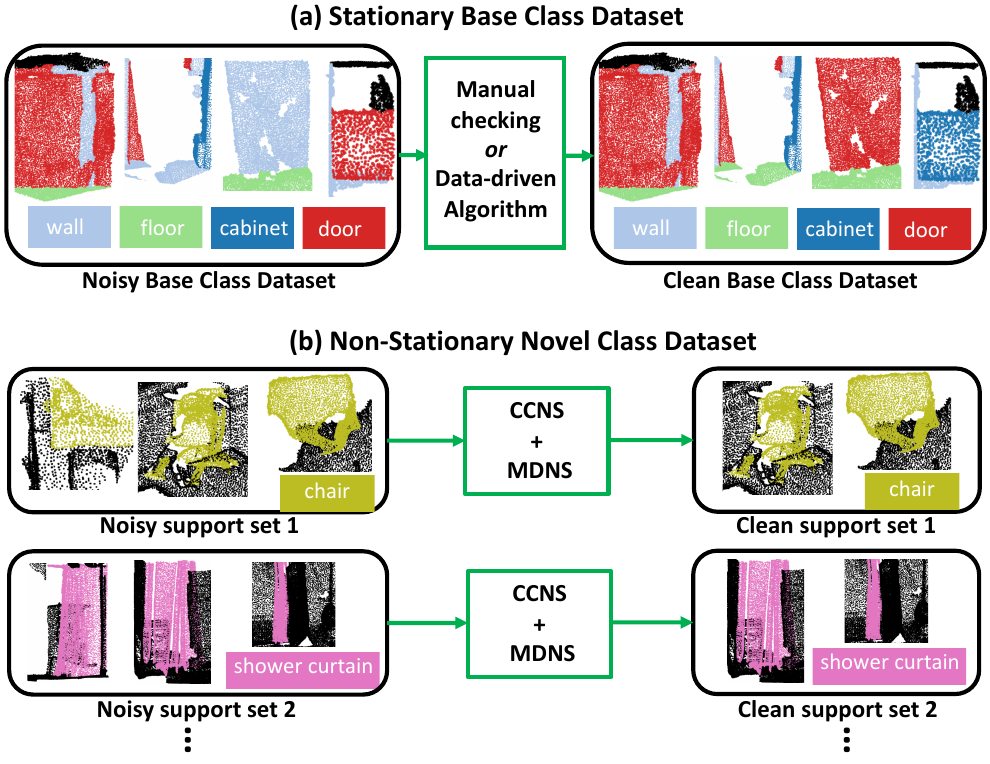}
\vspace{-3mm}
\caption{
\small{Comparison between noisy base and novel class dataset of 
3DFSSeg. (a) Base class dataset is static with finite samples. (b) Novel class dataset is non-stationary as new classes are continuously collected in the online testing stage. 
An example where a sofa and a curtain are wrongly annotated in support set 1 and 2, respectively.
}
}
\label{teaser}
\vspace{-2mm}
\end{wrapfigure}

As shown in Fig.~\ref{teaser}, we can refine the noisy annotations of the static base class dataset offline by either manual checking or data-driven algorithm \cite{ye2021learning} given enough time and budget.
However, it is impossible to invest the same amount of human supervision to guarantee noise-free in every support set after model being deployed because the number of new classes in the real world is \textit{infinite} \cite{ purushwalkam2022challenges,hariharan2017low}. 
Neither can we use data-driven algorithm \cite{ye2021learning} to automatically clean the noise due to severe overfitting to the small number of training samples per new class (\cf Tab.~\ref{s3dis-main}).

To this end, we tackle with the noisy labels in the testing stage of 3DFSSeg, which is challenging but with high practical value.
In 3DFSSeg, a few support point clouds are provided as learning samples for each new class during meta-testing. Each support sample (\ie shot) is provided with a binary mask indicating the presence of the corresponding class. Based on the given support set, the model segments the new class in any unlabeled (\ie query) point clouds.
As pointed out by \cite{ye2021learning, humblot2023cad} that the instance-level noise is most common in the annotation, objects of other classes are wrongly annotated as the target class and collected in the support set.
We define shots with incorrectly labeled foreground object as noisy shots. Thus, the goal of robust few-shot point cloud semantic segmentation (R3DFSSeg) is to learn a robust few-shot segmentor that is less influenced by the noisy shots.  



In this paper, we first propose a Component-level Clean Noise Separation (CCNS) representation learning to learn robust representation that is discriminative between features of clean and noisy points.  
Inspired by \cite{zhao2021few}, we adopt the meta-learning paradigm for few-shot point cloud segmentation. 
During meta-training, we randomly inject noise into the support set by sampling point clouds containing foreground objects from other classes to mimic the noisy meta-testing environments. 
We introduce a class-wise supervised contrastive learning on the noisy support set to separate the clean samples of the target classes from the noisy samples. To obtain more fine-grained and diverse contrastive features, we further propose the use of farthest point sampling to decompose the masked points in the feature space into multiple components. Intuitively, our CCNS is designed to encourage features from different classes to be well-separated, 
such that the clean shots in the support set would form the largest cluster in the feature space when learning converges.


We further propose a Multi-scale Degree-based Noise Suppression (MDNS) scheme to remove the noisy shots from the support set during testing stage. Our MDNS separates clean from noisy samples by checking the degree of each sample in a fully connected pair-wise similarity graph. Clean samples tend to form well-defined clusters with higher degrees in the pair-wise similarity graph. In contrast, noisy samples are relatively scattered with lower degrees of connectivity in the feature space.

Our \textbf{main contributions} can be summarized as follows: \textbf{1)} To the best of our knowledge, we are the first to study the problem of robust few-shot point cloud semantic segmentation, which is important in real-world applications since noisy labels are inevitable in practice.
\textbf{2)} We propose a component-level clean noise separation method for representation learning to enhance the class-level discrimination in the embedding space. 
\textbf{3)} We propose a multi-scale degree-based noise suppression scheme that is able to effectively remove noisy samples from the small support set for each new class during testing.
\textbf{4)} We conduct extensive experiments on two benchmark datasets (\ie S3DIS and ScanNet) with various noise settings and show superior results over the baselines.

\vspace{-5mm}
\section{Related Work}
\vspace{-3mm}
\paragraph{Few-shot Learning.}
Few-shot learning aims to transfer knowledge learned from the abundant samples of the seen class to a set of unseen classes with only few labeled samples. One of the dominant approach is the metric-based \cite{snell2017prototypical, vinyals2016matching} methods, which meta-learns a transferable feature embedding that coincides with a fixed metric. The pioneer work ProtoNet \cite{snell2017prototypical} predicts query label by finding the nearest class prototype under the Euclidean distance. 
The key to the metric-based method is the discriminative feature embedding with compact class clusters \cite{ma2021partner,yang2022few, gao2021contrastive, ye2020few}. 
Ye \etal \cite{ye2020few} apply the contrastive objective to align the training instances close to its own class center after the embedding adaptation. 
Although we also use contrastive learning in the episodic training, we adopt fine-grained contrastive objective (\ie feature components) to better capture the diverse intra-class distribution of point cloud.

\vspace{-6mm}
\paragraph{Few-shot Semantic Segmentation.}
Few-shot semantic segmentation segments semantic objects in an image \cite{wang2019panet,zhang2019canet,liu2020part} or a point cloud \cite{zhao2021few, mao2022bidirectional, lai2022tackling} with only few annotated samples.
The 2D image semantic segmentation can be categorized into relation-based method \cite{zhang2019canet,zhang2019pyramid,tian2020prior, zhang2021self} and prototype-based method \cite{liu2020part, yang2020prototype, wang2019panet}. 
Zhao \etal \cite{zhao2021few} propose the first work on 3D few-shot point cloud semantic segmentation. They generate multi-prototypes via farthest point sampling to better capture the complex data distribution of the point cloud. The transductive inference is conducted between multi-prototypes and query points to infer the label for each query point. However, all these works assume that the annotation in the given support are accurate during testing time. In practice, this is a very strong assumption given that the pixel-level and point-level annotation are extremely tedious and error-prone. In view of this limitation, this paper studies the problem of robust few-shot point cloud semantic segmentation and proposes a effective model that can better adapt to real world applications.

\vspace{-6mm}
\paragraph{Learning with Noisy Labels.}
Learning with noisy labels is gaining increasing attention as the deep neural networks are shown to be extremely vulnerable to the noisy labels \cite{arpit2017closer,han2018co,li2020dividemix}. There are three major approaches: label correction using the prediction of the model as the new label \cite{li2020dividemix, reed2014training, liu2020early, song2019selfie}, sample selection using small loss criterion to selectively update model \cite{han2018co,wei2020combating,yu2019does} and learning robust representation \cite{li2022selective,wu2021ngc,li2021learning,iscen2022learning,ortego2021multi}. 
PNAL \cite{ye2021learning} proposes robust point cloud semantic segmentation (R3DSeg) that studies label noise in the fully-supervised setting.
It stores the prediction history of every point in the training dataset and corrects point labels in the cluster-wise manner epoch by epoch. 
All these noise-robust methods deal with static dataset in the offline training stage and require massive samples to train (data-driven). In contrast, R3DFSSeg addresses noise in the online testing stage, where new classes appear continuously with small support set. 
The data-driven algorithms would thus overfit in R3DFSSeg. 


Existing methods dealing with noisy label of novel classes in the few-shot learning are only for 2D image classification (R2DFSL) \cite{liang2022few,mazumder2021rnnp,lu2020robust}.  RNNP \cite{mazumder2021rnnp} proposes a non-parametric test method by combining data augmentation with k-means to refine the class prototype. 
Liang \etal \cite{liang2022few} learn a robust prototype generater by relying on the self-attention module inside the Transformer \cite{vaswani2017attention} to weigh down the noisy shots. 
Compared to 2D classification, 3D point cloud segmentation is more challenging as it requires per-point classification and point cloud has much larger intra-class variance. Thus, the 2D methods, which only generate one robust prototype per class, fail on the R3DFSSeg.


\begin{figure*}[t]
\centering
\includegraphics[scale=0.4]{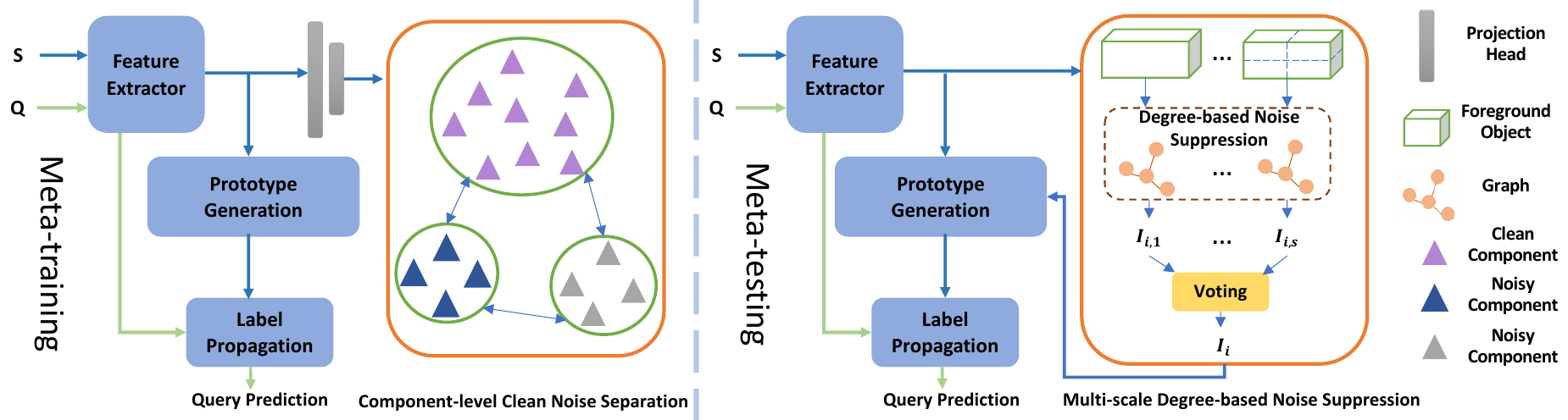}
\vspace{-2mm}
\caption{\small{\textbf{The architecture of our framework}. `S' represents the support point cloud and `Q' represents the query point cloud. The left figure shows the pipeline during meta-training, where we conduct component-level clean noise separation representation learning for each episode class. Components of different classes are pushed away from each other. The right figure shows the pipeline during meta-testing, where we perform multi-scale degree-based noise suppression to remove the noisy shots.}}
\label{framework}
\vspace{-5mm}
\end{figure*}

\vspace{-5mm}
\section{Our Method}
\label{sec:formatting}

\vspace{-3mm}
\paragraph{Problem Formulation.}
The few-shot point cloud segmentation consists of two datasets: $\mathcal{T}_{base}$ and $\mathcal{T}_{novel}$ sampled from disjoint classes $\mathcal{C}_{base}$ and $\mathcal{C}_{novel}$, respectively. The goal is to learn a model from $\mathcal{C}_{base}$ and generalize to the $\mathcal{C}_{novel}$. Following previous work \cite{zhao2021few}, we adopt the episodic training on the $\mathcal{C}_{base}$ to emulate the few-shot setting during testing.
In each $N$-way $K$-shot episode, $N$ is the number of classes to be learned, and $K$ is the number of labeled samples per class. The labeled samples are termed as the support set:  $S=\left\{\left(P_k^1, M_k^1\right)_{k=1}^K, \ldots,\left(P_k^N, M_k^N\right)_{k=1}^K\right\}$. Each point cloud $P_k^n \in \mathbb{R}^{m \times f_0}$ contains $m$ points with input feature dimension of $f_0$. The $M_k^n \in \mathbb{R}^{m \times 1}$ is the corresponding binary mask indicating the presence of class $n$.


We are also given a set of $T$ unlabeled point clouds, termed as the query set: $Q=\left\{\left(R_i, L_i\right)\right\}_{i=1}^T$. Each query point cloud $R_i \in \mathbb{R}^{m \times f_0}$ 
is associated with the ground truth label $L_i \in \mathbb{R}^{m \times 1}$ only available in the training stage. 
During testing, $M_k^n$ can wrongly assign object of another class to
class $n$ due to the instance-level labeling error \cite{ye2021learning}.
We denote the noisy mask $\Tilde{M}_k^n$ and the corresponding point cloud $\Tilde{P}_k^n$ as the noisy sample, and its correct class assignment as $Y_k$. 
Consequently, the support set $S$ becomes the mixture of clean and noisy shots. The goal of robust few-shot point cloud semantic segmentation is to correctly predict the query label by learning from the noisy support set $S$.

\vspace{-4mm}
\paragraph{Framework Overview.}
Fig.~\ref{framework} illustrates our proposed framework. We choose AttMPTI \cite{zhao2021few} as our few-shot segmentor since it achieves state-of-the-art performance in the few-shot point cloud segmentation. 
In addition, AttMPTI is potentially robust to the noise when a good feature embedding is guaranteed (Sec.~\ref{why choose attmpti}).    
In view of this, we propose the Component-level Clean Noise Separation (CCNS) representation learning during meta-training to enhance the discrimination and generalization of the feature embedding for AttMPTI (Sec.~\ref{ccns}).
We further propose the multi-scale degree-based noise suppression (MDNS) to remove the noisy shots during meta-testing based on their similarity graph (Sec.~\ref{mdns}).

\subsection{Why Choose AttMPTI?}
\label{why choose attmpti}
\begin{wrapfigure}{r}{0.48\textwidth}
\vspace{-2mm}
\centering
\includegraphics[scale=0.38]{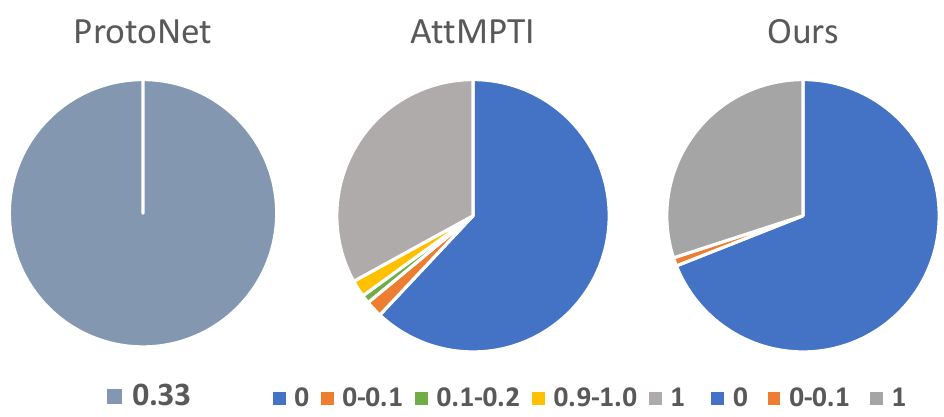}
\vspace{-3mm}
\caption{\small{Comparison of prototype cleanness from different methods on a 5-shot with 40\% out-episode noise setting. `1' means the prototype only containing clean-labeled points, and `0' means the prototype only containing points that are incorrectly labeled as the target class. Values in between 0-1 represent the portion of clean-labeled points in the prototype.}}
\label{proto-cleanness}
\vspace{-3mm}
\end{wrapfigure}

AttMPTI \cite{zhao2021few} is the state-of-the-art few-shot point cloud segmentation method. It consists of a feature extractor to embed the support and query point cloud into the same metric space, a multi-prototype generation module to generate prototypes from support set, and a label propagation module to infer query label. 
Compared to 
ProtoNet \cite{snell2017prototypical}, AttMPTI has several unique components that gives it the potential to be robust, in addition to
showing more superior performance.
\textbf{First}, AttMPTI generates multi-prototypes via FPS \cite{qi2017pointnet++}, while ProtoNet uses mean aggregation of all the relevant class feature. The sampled seed points via FPS are able to represent the diversity of the feature space, and the local prototype is generated by clustering each point to the nearest seed point based on the Euclidean distance in the feature space. In this way, the multi-prototypes can inherently separate the clean and noisy points in the prototype-level. As shown in Fig.~\ref{proto-cleanness}, the clean ratio of local prototypes is either 1 (100\% clean) or 0 (100\% noise), but it seldom produces a half-clean prototype. 
In comparison, the global prototype used in the ProtoNet leads to a clean-noise compound.
\textbf{Second}, AttMPTI infers query labels via label propagation \cite{zhou2003learning} in a transductive fashion, while ProtoNet infers each query point independently with the set of class prototypes. The label propagation is based on the manifold smoothness, \ie nearby samples in the feature space share the same label, and it has the ability to correct the noisy label \cite{wu2021ngc, iscen2022learning}. In contrast, 
ProtoNet independently and identically predicts the label for each query point based on the global prototypes that are potentially noisy.
The lack of reasoning the relationships among the support and query prevents the model from being able to correct the support noise. 
Although the design of AttMPTI shows a better potential than ProtoNet in resisting the noise existing in the support set, 
the performance of both multi-prototype generation and label propagation are subjected to the discriminativity of the feature embeddings. To enhance the representation learning, we propose to perform component-level clean-noise separation.

\begin{wrapfigure}{R}{0.48\textwidth}
\vspace{-2mm}
\centering
\includegraphics[scale=0.45]{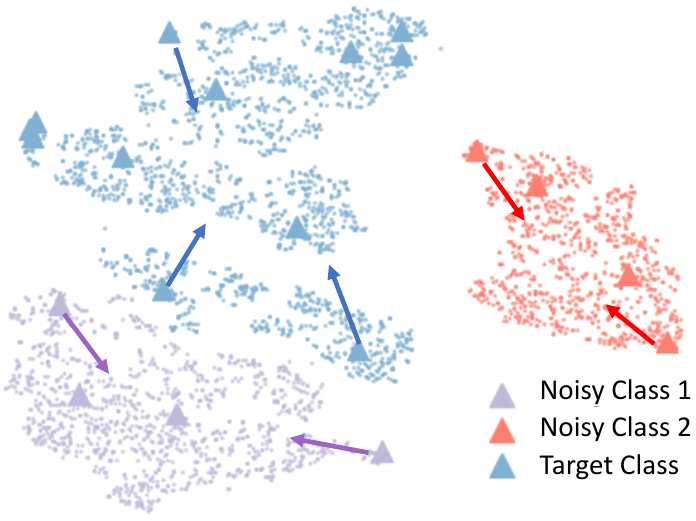}
\vspace{-3mm}
\caption{\small{t-SNE \cite{van2008visualizing} visualization of the CCNS on a 5-shot support set with 2 noisy shots. Each dot represents a point in the feature space and each triangle represents a feature component. Different colors represent different classes with blue indicating the target class. The arrow shows the direction to pull the feature components.}}
\label{component}
\vspace{-1mm}
\end{wrapfigure}

\subsection{Component-level Clean Noise Separation}
\label{ccns}
Our component-level clean noise separation (CCNS) representation learning aims to enhance the class-wise discrimination in the feature space.
We randomly replace some of the K support shots with shots sampled from other classes during episodic training and induce the model to differentiate clean and noisy shots in the feature space.
With these synthesized support sets with noisy labels, we perform a clean-noise separation representation learning for each way (\ie class) by optimizing the model with the class-wise contrastive learning among the $K$ support shots as follow:
\begin{equation}
\small
\mathcal{L}_{\text{CNS}}=
\frac{1}{K}\sum_{k=1}^K \left( \frac{-1}{|A(z_{k})|} \sum_{z_{g} \in A(z_{k})} \log \frac{\exp \left(z_{k} \cdot z_{g} / \tau\right)}{\sum\limits_{h \backslash k} \exp \left(z_{k} \cdot z_{h} / \tau\right)} \right),
\end{equation}
where $z_{k} \in \mathbb{R}^{d}$ is the L2 normalized average foreground feature of the support point cloud $P_k$ in the projection space. 
$A(z_{k}) = \left\{z_{g} \mid Y_{g} = Y_{k}\right\}$ is the set of positive samples $z_g$ with its semantic label $Y_g$ the same as the semantic label $Y_k$ of $z_k$. $|A(z_{k})|$ is the cardinality and $\tau$ is the temperature. By training with $\mathcal{L}_{\text{CNS}}$, the shots with same foreground class are encouraged to stay together while staying away from samples of other classes.


\ytComment{
Unfortunately, a simple mean aggregation of the foreground area tends to be sub-optimal in representing the class distribution since the distribution of point features of each class is very large as shown in Fig.~\ref{component}. 
To this end, we conduct class-wise contrastive learning in a more fine-grained way by dividing the features in each foreground area into local components. 
The feature components aggregate local patterns that exhibit similar fine-grained semantics, and have better coverage of the feature space compared to the 
naive mean aggregation.
}
Specifically, we first perform FPS in the feature space and then locally aggregate the point features into a set of feature components $\left\{z_{k}^1,\cdots, z_{k}^R \right\}$, to replace the original holistic $z_{k}$.
Consequently, the component-level clean noise separation $\mathcal{L}_{\text{CCNS}}$ is formulated as:

\vspace{-3mm}
\begin{equation}
\mathcal{L}_{\text{CCNS}} = \frac{1}{KR}\sum\limits_{k=1}^K \sum\limits_{i=1}^R \left( \frac{-1}{|A(z_{k}^i)|} \sum\limits_{z_{g}^j \in A(z_{k}^i)} \log \frac{\exp\left(z_{k}^i \cdot z_{g}^j / \tau\right)}{\sum\limits_{h,b \backslash (k,i)} \exp\left(z_{k}^i \cdot z_{h}^b / \tau\right)} \right),
\end{equation}
%
where the $A(z_{k}^i) = \left\{z_{g}^j \mid Y_{g} = Y_{k}\right\}$ is the set of positive samples with the same semantic label $Y_g$ as $Y_k$, and the $|A(z_{k}^i)|$ is the cardinality. 
As shown in Fig.~\ref{component}, each component represents a different aspect of its corresponding shot in the feature space. 
Essentially, it forms a multi-view self-supervised contrastive learning for each shot, where the `view' is a local component in the feature space. 
Correspondingly, the components at the boarder of the class distribution automatically serve as the hard negative samples to other classes and hard positive samples to its own class, which are the key to a successful contrastive learning \cite{khosla2020supervised,chen2020simple}.

The final optimization objective during the training stage is given by:

\vspace{-3mm}
\begin{equation}
\mathcal{L} = \mathcal{L}_{\text{CE}} + \lambda \mathcal{L}_{\text{CCNS}},
\end{equation}
%
where $\lambda$ is a hyper-parameter to weigh the contribution of $\mathcal{L}_{\text{CCNS}}$. $\mathcal{L}_{CE}$ is the original cross-entropy loss in AttMPTI.

\vspace{-3mm}
\subsection{Multi-scale Degree-based Noise Suppression}
\label{mdns}
Although the clean and noisy points can separate under the well-learned embedding space, the prototype generation and label propagation module are still exposed to the mislabeled shots during testing time. To reduce their negative influence during testing, we design a degree-based noise suppression scheme to automatically remove the suspicious noisy shots. Specifically, we build a fully connected graph G on the K support shots for each way.
We average the foreground feature $x_i \in \mathbb{R}^d$ of the $i$-th shot as the feature of node i. 
The weight $W_{ij}$ of the edge encodes the affinity between the two end nodes $i$ and $j$ as follow:
\vspace{-3mm}
\begin{equation}
W_{i j}:= \begin{cases}{\left[x_i^{\top} x_j\right]_{+}^\gamma,} & \text { if } i \neq j \\ 0, & \text { otherwise }\end{cases}.
\vspace{-1mm}
\end{equation}
We then compute the degree $d_i = \sum_j W_{ij}$ for each node i. Essentially, the degree reflects the nodes connection in the graph. 
The noisy shots tend to have lower degree since the clean shots usually form a cluster with the largest size and the noisy shots are scattered in the feature space.
Consequently, we identify them based on the clean indicator:
\vspace{-3mm}
\begin{equation}
I_i:= \begin{cases} 1 & \text { if } d_i > thr \\ 0, & \text { otherwise }\end{cases},
\vspace{-2mm}
\end{equation}
where we set the $thr$ as the mean of the $\left\{d_i \right\}_{i=1}^K$. The shots with $I=0$ are treated as noise and removed. 

Some point clouds may have complex data distribution that cannot be sufficiently represented by a global representation. To mitigate this problem, we extend the single-level degree-based noise suppression scheme to multi-level, thus yielding the Multi-scale Degree-based Noise Suppression (MDNS).  
Our MDNS can be more robust to some complex samples and consequently improve the accuracy of clean sample identification. 
Specifically, we add an additional level to perform noise suppression. 
We evenly split the foreground object along the x/y/z coordinates, and denote the number of cuts along the x/y/z coordinates as $n_x$/$n_y$/$n_z$. The foreground feature in each sub-shot is locally aggregated and the feature set for each shot is enlarged to $\left\{x_{i,s}^1, \cdots, x_{i,s}^e\right\}$, where $e=n_x\times n_y\times n_z$. The single representation $x_i$ is the case of $\left\{n_x=1, n_y=1, n_z=1 \right\}$ and is considered as the coarsest scale with $s=1$. 
We then send them into the noise suppression module to get the clean indicator $\left\{I_{i,s}^1, \cdots, I_{i,s}^e\right\}$, where the majority voting is performed get the shot-level indicator $I_{i,s}$. Lastly, we assemble the final prediction $I_i$ as the majority voting of the prediction at each scale $\left\{I_{i,1}, \ldots, I_{i,s} \right\}$.

\vspace{-4mm}
\section{Experiments}

\subsection{Datasets and Noise Settings}
\paragraph{Datasets.}
We conduct experiments on \textbf{S3DIS} \cite{armeni20163d} and \textbf{ScanNet} \cite{dai2017scannet}. S3DIS contains point clouds of 272 rooms collected from six indoor areas with annotation of 12 semantic classes. ScanNet contains point clouds of 1,513 scans from 707 unique indoor scenes with annotation of 20 semantic classes. Following \cite{zhao2021few}, we split each room into non-overlapping blocks with size of $1\text{m} \times 1\text{m}$ on the xy plane. Consequently, S3DIS and ScanNet contains 7,547 and 36,350 blocks, respectively. We sample $m=2,048$ points as the input point cloud from a block. The input feature $f_0$ corresponds to XYZ, RGB and normalized XYZ values.
During training, we randomly sample one episode by first sampling N classes from $\mathcal{C}_{base}$ and then sampling $NK$ point clouds as the support set and $T$ point clouds as the query set. The support mask $M$ and the query label $L$ are modified from its original annotation to only indicate the presence of the target classes with irrelevant classes as the background. The testing episodes are formed in a similar way, except for that we exhaustively sample 100 episodes for each combination of N classes from the $\mathcal{C}_{novel}$. We use the data split 0 of \cite{zhao2021few} as the test classes on both datasets. We adopt the mean Intersection over Union (mIoU) as the evaluation metric.

\begin{table*}[t]
\centering
\resizebox{0.95\linewidth}{!}{
\begin{tabular}{p{2cm}<\centering|p{1.5cm}<\centering p{1.5cm}<\centering|p{1.5cm}<\centering p{1.5cm}<\centering p{1.5cm}<\centering p{1.5cm}<\centering |p{1.5cm}<\centering p{1.5cm}<\centering p{1.5cm}<\centering p{1.5cm}<\centering}
\hline
\multirow{3}{*}{model} & \multicolumn{2}{c|}{\multirow{2}{*}{0\%}} & \multicolumn{4}{c|}{In-episode Noise}                            & \multicolumn{4}{c}{Out-episode Noise}                           \\ \cline{4-11} 
                       & \multicolumn{2}{c|}{}       & \multicolumn{2}{c|}{20\%}            & \multicolumn{2}{c|}{40\%} & \multicolumn{2}{c|}{40\%}            & \multicolumn{2}{c}{60\%} \\ \cline{2-11} 
                       & 2-way                       & 3-way       & 2-way  & \multicolumn{1}{c|}{3-way}  & 2-way       & 3-way       & 2-way  & \multicolumn{1}{c|}{3-way}  & 2-way       & 3-way      \\ \hline
                       
PNAL \cite{ye2021learning}  &13.67 &8.12  &8.94 &\multicolumn{1}{c|}{5.45}   &5.95 &3.13   &8.08 &\multicolumn{1}{c|}{4.28}    &4.77    &2.87     \\
Tra-NFS \cite{liang2022few}  & 44.98   &31.67        & 43.44   &\multicolumn{1}{c|}{30.68}         &37.27     & 27.39    & 41.72     &\multicolumn{1}{c|}{28.43}   &35.67   &23.20    \\
ProtoNet \cite{snell2017prototypical}               & 57.02                      & 46.78      & 54.21 & \multicolumn{1}{c|}{43.57} & 42.57      & 36.71      & 50.01 & \multicolumn{1}{c|}{39.31} & 44.96      & 36.08     \\
AttMPTI \cite{zhao2021few}                & 65.90                      & 51.71      & 60.01 & \multicolumn{1}{c|}{47.96} & 38.81      & 37.56        & 58.60 & \multicolumn{1}{c|}{44.76}   & 51.18      & 40.32     \\

\textbf{Ours}                   & \textbf{68.21}                      & \textbf{54.79}      & \textbf{66.02} & \multicolumn{1}{c|}{\textbf{52.91}} & \textbf{58.01}      & \textbf{48.72}      & \textbf{66.09} & \multicolumn{1}{c|}{\textbf{50.71}} & \textbf{58.84}      & \textbf{46.19}     \\ \hline
\end{tabular}
}
\caption{\small{Results on the S3DIS using mIoU metric on 2-way 5-shot and 3-way 5-shot.}}
\label{s3dis-main}
\vspace{-3mm}
\end{table*}

\begin{table*}[t]
\centering
\resizebox{0.95\linewidth}{!}{
\begin{tabular}{p{2cm}<\centering|p{1.5cm}<\centering p{1.5cm}<\centering|p{1.5cm}<\centering p{1.5cm}<\centering p{1.5cm}<\centering p{1.5cm}<\centering |p{1.5cm}<\centering p{1.5cm}<\centering p{1.5cm}<\centering p{1.5cm}<\centering}
\hline
\multirow{3}{*}{model} & \multicolumn{2}{c|}{\multirow{2}{*}{0\%}} & \multicolumn{4}{c|}{In-episode Noise}                            & \multicolumn{4}{c}{Out-episode Noise}                           \\ \cline{4-11} 
                       & \multicolumn{2}{c|}{}       & \multicolumn{2}{c|}{20\%}            & \multicolumn{2}{c|}{40\%} & \multicolumn{2}{c|}{40\%}            & \multicolumn{2}{c}{60\%} \\ \cline{2-11} 
                       & 2-way                       & 3-way       & 2-way  & \multicolumn{1}{c|}{3-way}  & 2-way       & 3-way       & 2-way  & \multicolumn{1}{c|}{3-way}  & 2-way       & 3-way      \\ \hline
Tra-NFS \cite{liang2022few} & 41.89  & 31.56   & 39.72  &\multicolumn{1}{c|}{29.20}   & 34.25  &25.07  & 38.42  &\multicolumn{1}{c|}{27.29}   & 34.68   &23.78 \\
ProtoNet \cite{snell2017prototypical}               & 47.55                      & 38.97      & 44.19 & \multicolumn{1}{c|}{36.46} & 34.57      & 30.23      & 42.47 & \multicolumn{1}{c|}{33.88} & 36.64      & 28.55     \\
AttMPTI \cite{zhao2021few}                & \textbf{54.16}                      & \textbf{44.52}      & 46.63 & \multicolumn{1}{c|}{38.83} & 31.57      & 27.62      & 43.31 & \multicolumn{1}{c|}{34.33} & 36.45      & 26.79     \\
\textbf{Ours}                   & 53.50                            & 43.84      & \textbf{49.78} & \multicolumn{1}{c|}{\textbf{41.01}} & \textbf{38.70}      & \textbf{34.03}      & \textbf{47.90} & \multicolumn{1}{c|}{\textbf{38.93}} & \textbf{38.42}      & \textbf{28.81}     \\ \hline
\end{tabular}
}
\caption{\small{Results on the ScanNet using mIoU metric on 2-way 5-shot and 3-way 5-shot.}}
\label{scannet-main}
\vspace{-6mm}
\end{table*}

\paragraph{Noise Settings.}
We explore two types of label noise: 1) \textbf{In-episode noise} samples noisy shots from  other N-1 classes of the current episode. It studies how the mix of the N foreground classes affects the prediction of query point. We test the models on in-episode noise ratio of 20\% and 40\%. 2) \textbf{Out-episode noise} samples noisy shots from outside of the N classes in the $\mathcal{C}_{novel}$. It studies how the outliers affect the prediction of the query point. We test the models on out-episode noise ratio of 40\% and 60\%. 

The noise rate is defined as the percentage of the $K$ support shots. Following existing literature of learning with noisy labels \cite{han2018co, li2020dividemix, liang2022few}, we define the noise ratio with the restriction that the percentage of clean labeled samples is larger than any noisy class. We thus can only consider up to 40\% noise for the in-episode noise and up to 60\% noise for the out-episode noise in both 2-way 5-shot and 3-way 5-shot point cloud segmentation.

\vspace{-3mm}
\subsection{Implementation Details}
We adopt the AttMPTI \cite{zhao2021few} as the few-shot segmentor and follow the same training procedure as AttMPTI. We first pre-train the feature extractor 100 epochs on the $\mathcal{C}_{base}$ with learning rate of 0.001 and Adam optimizer. In the episodic training, the feature extractor is fine-tuned with learning rate of 0.0001 and other learnable modules are optimized with learning rate of 0.001. The projection head consists of one fully-connected layer with the output dimension d as 128. Both $\lambda$ and $\tau$ are set to 0.1, and R is set 4 in $\mathcal{L}_{\text{CCNS}}$. We randomly generate noisy shots in each episode during training by sampling shots from $\mathcal{C}_{base}$. The noise ratio is randomly chosen from $\left\{0,0.2,0.4\right\}$. In MDNS, the $\gamma$ is set to 3 for scale s=1 and to 1 for any other scale.
The MDNS is conducted in two scales: $\left\{n_x=1, n_y=1, n_z=1 \right\}$ and $\left\{n_x=2, n_y=2, n_z=1 \right\}$. All the experiments are done using one GTX 3090 GPU.


\begin{figure*}[t]
\centering
\includegraphics[scale=0.38]{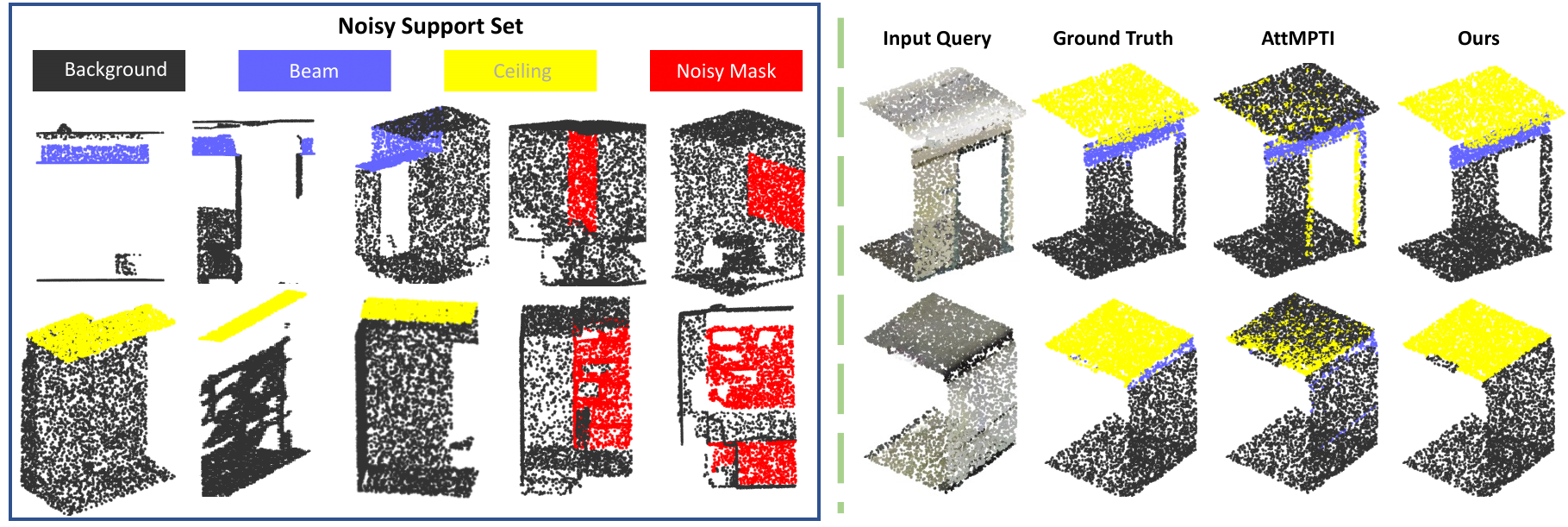}
\vspace{-3mm}
\caption{\small{Qualitative comparison of a 2-way 5-shot point cloud segmentation with 40\% out-episode noise on S3DIS.}}
\label{s3dis-seg}
\vspace{-2mm}
\end{figure*}

\vspace{-3mm}
\subsection{Main Results}
We compare with 3DFSSeg methods AttMPTI \cite{zhao2021few} and ProtoNet \cite{snell2017prototypical}, R2DFSL method Tra-NFS \cite{liang2022few} and R3DSeg method PNAL \cite{ye2021learning}. All methods use the same feature extractor as AttMPTI for fair comparison.

Tab.~\ref{s3dis-main} and Tab.~\ref{scannet-main} presents the experiment results on the noisy 2-way 5-shot and 3-way 5-shot point cloud segmentation on S3DIS and ScanNet, respectively. 
AttMPTI \cite{zhao2021few} usually has better performance than ProtoNet \cite{snell2017prototypical} in terms of various noise setting, but both 
methods suffer 
considerably with increasing noise ratio. It suggests that few-shot segmentor is indeed vulnerable to the support noise. On the other hand, our method is able to largely improve the robustness of the AttMPTI over all noise settings on both datasets.
Fig.~\ref{s3dis-seg} presents the qualitative results of a 2-way 5-shot point cloud segmentation with 40\% out-episode noise on the S3DIS. It shows our method can correctly segments the target classes in the query point while AttMPTI fails. 
\ytComment{
We notice that our model is slightly worse 
than AttMPTI in the 0\% setting in Tab.~\ref{scannet-main}. We postulate 
that our method can predict correct labels, but the noisy ground truths of ScanNet \cite{ye2021learning} cannot reflect the true performance of our method. This postulation is evidenced by the great superiority of our method over baseline methods on S3DIS, which is a dataset with clean ground truths. 
It suggests that our method can adapt to the unknown test environment (both clean and noise test), which is important for model deployment in real world.
}



2D robust few-shot learner Tra-NFS \cite{liang2022few} performs poorly on R3DFSSeg due to severe modality gap, \ie point cloud has larger intra-class viriance than 2D images, making Tra-NFS hard to detect clean shots. 3D robust point cloud segmentor PNAL \cite{ye2021learning} also fails in the few-shot setting due to small support set in each episode.

We further notice that the in-episode noise has larger negative influence than the out-episode noise, \eg 40\% in-episode noise vs 40\% out-episode noise. 
We believe the reason is that the features in each foreground class usually form a compact cluster. The in-episode noise causes the labels in this compact cluster to be different, which severely confuses the model of which class this cluster belongs to. In contrast, the out-episode noise are usually separated from the foregrond classes in the feature space, and is less likely to influence them.

\begin{wraptable}{R}{0.45\textwidth}
\centering
\vspace{-4mm}
\resizebox{\linewidth}{!}{
\begin{tabular}{c|c|cc|cc}
\hline
\multirow{2}{*}{model} & \multirow{2}{*}{0\%} & \multicolumn{2}{c|}{In-episode Noise} & \multicolumn{2}{c}{Out-episode Noise} \\ \cline{3-6} 
                       &                      & 20\%              & 40\%              & 40\%              & 60\%              \\ 
\hline
AttMPTI                & 65.90                     & 60.01            &38.81                  &58.60                   & 51.11        \\ 
AttMPTI+CCNS          & \textbf{68.50}             &63.10             &41.75                  & 63.77                  & 56.79   \\
AttMPTI+MDNS          &  64.80                     & 63.03           & 52.78                & 61.73                 & 52.98   \\
\textbf{Ours}                    & 68.21                     & \textbf{66.02}             &\textbf{58.01}                 &\textbf{66.09}                  &\textbf{58.84}   \\
\hline
\end{tabular}
}
\vspace{-3mm}
\caption{\small{Effectiveness of CCNS and MDNS on the S3DIS on 2-way 5-shot. `Ours' consists of both CCNS and MDNS.}}
\label{s3dis-cl+cd}
\vspace{-4mm}
\end{wraptable}

\vspace{-4mm}
\subsection{Ablation Study}
\paragraph{Effectiveness of CCNS and MDNS.}
We analyze the effectiveness of our proposed component-level clean noise suppression (CCNS) and multi-scale degree-based noise suppression (MDNS) on S3DIS in Tab.~\ref{s3dis-cl+cd}.
It is worth noting that the robustness of AttMPTI is improved by simplely adding our feature representation learning, \ie CCNS. It verifies our claim that AttMPTI has the potential to be noise robust (by FPS based multi-prototype generation and label propagation), yet is subject to how discriminative the feature embedding is. 
\ytComment{
MDNS improves performance on most settings and its performance gains are also subjected to the discriminativity of the feature space. 
By further equipping with CCNS, our final model achieves consistent and significant improvements under all settings.
}
The ablation study on hyperparameter choosing is provided in the \textbf{supplementary material}. 

\begin{wraptable}{R}{0.45\textwidth}
\centering
\vspace{-1mm}
\resizebox{\linewidth}{!}{
\begin{tabular}{c|c|cc|cc}
\hline
\multirow{2}{*}{model} & \multirow{2}{*}{0\%} & \multicolumn{2}{c|}{In-episode Noise} & \multicolumn{2}{c}{Out-episode Noise} \\ \cline{3-6} 
                       &                      & 20\%              & 40\%              & 40\%              & 60\%              \\ 
\hline
AttMPTI                & 32.75                    & 27.96            &20.72                 &23.89                   & 17.54        \\ 
\textbf{Ours} & 32.74  &\textbf{30.79} 	&\textbf{26.73}	 & \textbf{28.13}	&\textbf{21.22} \\
\hline
\end{tabular}
}
\vspace{-3mm}
\caption{\small{5-way 5-shot setting on ScanNet.}}
\label{5-way}
\vspace{-4mm}
\end{wraptable}

\paragraph{High way setting.}
Tab.~\ref{5-way} shows results of 5-way 5-shot setting on ScanNet. Our model again can significantly outperform AttMPTI on all noise settings.

\vspace{-4mm}
\section{Conclusion}
In this paper, we address the new task of robust few-shot point cloud segmentation, which is a more general setting that considers label noise in the support set. We design the Component-level Clean Noise Separation (CCNS) representation learning to learn a discriminative feature embedding. Our CCNS encourages the features from different classes to stay away from each other, and concurrently induces the clean shots to form the largest cluster in the feature space. Leveraging the clean samples identified from our CCNS, we further propose the Multi-scale Degree-based Noise Suppression (MDNS) to remove the noisy shots before the prototype generation based on their affinity with other samples in the support set. 
Experiment results that outperform the baselines show the feasibility of our proposed method. 

\paragraph{Acknowledgement.}This research is supported by the National Research Foundation, Singapore under its AI Singapore Programme (AISG Award No: AISG2-RP-2021-024),
and the Tier 2 grant MOE-T2EP20120-0011 from the Singapore Ministry of Education. This research is also supported by the SUTD-ZJU Thematic Research Grant RS-MEZJU-00031. The work is fully done at the National University of Singapore.

\appendix
\renewcommand\thefigure{\Alph{section}\arabic{figure}}  
\renewcommand\thetable{\Alph{section}\arabic{table}}

{\centering\section*{Supplementary Material}}

\section{Ablation Study}
\setcounter{figure}{0}
\setcounter{table}{0}
\paragraph{Analysis of different R values.}
Tab.~\ref{abalation+R} shows the ablation study of different number of components for each shot in the component-level clean noise separation. `R=1' is the shot-level representation. It can be seen that the performance of `R=1' is generally worse than that of the component-level contrastive learning, which verifies that the feature is sub-optimized with a single holistic aggregation. By dividing into local components, we can get more fine-grained and diverse positive and negative samples with `R=4' having the best performance.

\begin{table}[h]
\centering
\setlength{\abovecaptionskip}{0.1cm}
\resizebox{0.5\linewidth}{!}{
\begin{tabular}{c|c|cc|cc}
\toprule
\multirow{2}{*}{model} & \multirow{2}{*}{0\%}             & \multicolumn{2}{c|}{In-episode Noise} & \multicolumn{2}{c}{Out-episode Noise} \\ \cline{3-6} 
                       &                                  & 20\%              & 40\%              & 40\%              & 60\%              \\ \midrule
R=1                   &  67.62                                & 65.83                  & 55.94                  & 65.17                  &57.46                   \\
R=2            & 67.93                           &\textbf{66.28}             &57.41            & 65.08             & 58.57                 \\
R=4                    & \textbf{68.21}                           & 66.02            & \textbf{58.01}            & \textbf{66.09}            & \textbf{58.84}            \\
R=8                   &67.40                                 &65.58   &56.66                  & 65.52            &57.94                   \\ \hline
\end{tabular}
}
\caption{Effects of different number of components in CCNS.}
\vspace{-2mm}
\label{abalation+R}
\end{table}

\paragraph{Analysis of different noise ratios in CCNS.}
We analyze different combination of noise ratio in the episodic training since our component-level clean noise separation is conducted among the clean and noisy shots. 
`$\left\{0.2,0.4 \right\}$' has large performance drop when comparing with `$\left\{0,0.2,0.4 \right\}$', which suggests that it is very necessary to include noise-free episodes during training. By further adding noise ratio of 0.6 (with the restriction that any number of noisy class should not outnumber the clean shots), there is again a significant drop in performance. We can conclude that only a mix of a proper portion of noisy and clean episodes during training can bring decent improvement in the noisy test.

\begin{table}[h]
\centering
\setlength{\abovecaptionskip}{0.1cm}
\resizebox{0.5\linewidth}{!}{
\begin{tabular}{c|c|cc|cc}
\toprule
\multirow{2}{*}{Training Noise} & \multirow{2}{*}{0\%} & \multicolumn{2}{c|}{In-episode Noise} & \multicolumn{2}{c}{Out-episode Noise} \\ \cline{3-6} 
                       &                      & 20\%              & 40\%              & 40\%              & 60\%              \\ 
\midrule  
$\left\{0.2,0.4 \right\}$                &63.66                    &60.91            & 49.51                & 59.95                 & 50.55        \\ 
$\left\{0,0.2,0.4 \right\}$               & \textbf{68.21}                           & \textbf{66.02}            & \textbf{58.01}            & \textbf{66.09}            & \textbf{58.84}   \\
$\left\{0,0.2,0.4,0.6 \right\}$                   & 66.40                     & 65.39             & 55.00             & 63.39                &56.11   \\
\bottomrule
\end{tabular}
}
\caption{Effects of different simulated noise combinations.}
\vspace{-2mm}
\label{abalation_noisetrain}
\end{table}

\paragraph{Analysis of different scales in MDNS.}
Tab.~\ref{abalation_multiscale} presents the analysis of different scales in the multi-scale degree-based noise suppression. Due to space limitation, we only provide the comparison of selected scales from the many possibilities of combinations. We first analyze what constitutes a good scale. 
It is almost guaranteed that the holistic scale $\left\{1/1/1\right\}$ gives decent performance since the mean representation covers the general information. 
The performance varies a lot when the foreground objects are divided into fine-grained scales. By comparing $\left\{2/2/1\right\}$, $\left\{1/2/2\right\}$ and $\left\{2/1/2\right\}$, we can see that a cut on the z-axis causes a significant drop in performance on the heavy noise setting. 
By comparing $\left\{3/3/1\right\}$ with $\left\{2/2/1\right\}$, we can see that the cuts that are too fine-grained cause a performance drop due to the severe lack of the global information in the sub-shots. 
Overall, $\left\{1/1/1\right\}$ and $\left\{2/2/1\right\}$ are the good scales and their combination achieves the best performance.  

\begin{table}[h]
\centering
\setlength{\abovecaptionskip}{0.1cm}
\resizebox{0.5\linewidth}{!}{
\begin{tabular}{c|cc|cc}
\toprule
\multirow{2}{*}{$\left\{n_x/n_y/n_z\right\}$} & \multicolumn{2}{c|}{In-episode Noise} & \multicolumn{2}{c}{Out-episode Noise} \\ \cline{2-5} 
                                    & 20\%              & 40\%              & 40\%              & 60\%              \\ \midrule
$\left\{1/1/1 \right\}$                           & 65.94            & 57.27            & 65.90            & 58.70            \\
$\left\{2/2/1 \right\}$                          & \textbf{66.47}            & 56.56            & 65.96            & 57.99            \\
$\left\{1/2/2 \right\}$                         & 66.37            & 55.29            & 65.96            & 54.56            \\
$\left\{2/1/2 \right\}$                         & 66.31            & 54.81            & 65.44            & 54.96            \\
$\left\{3/3/1 \right\}$                          & 65.99            & 54.69            & 65.24            & 57.69            \\
$\left\{1/1/1 \right\}$ \& $\left\{2/2/1 \right\}$             & 66.02            & \textbf{58.01}            & \textbf{66.09}            & \textbf{58.84}            \\
$\left\{1/1/1 \right\}$ \& $\left\{2/2/1 \right\}$ \& $\left\{3/3/1 \right\}$            &65.81           &\textbf{58.01}           & 66.00            & 57.51           \\

\bottomrule
\end{tabular}
}
\caption{Effects of different scale choices in MDNS.}
\vspace{-3mm}
\label{abalation_multiscale}
\end{table}

\section{t-SNE Visualization}
\begin{figure}[h]
\centering
\setlength{\abovecaptionskip}{0.1cm}
\includegraphics[scale=0.48]{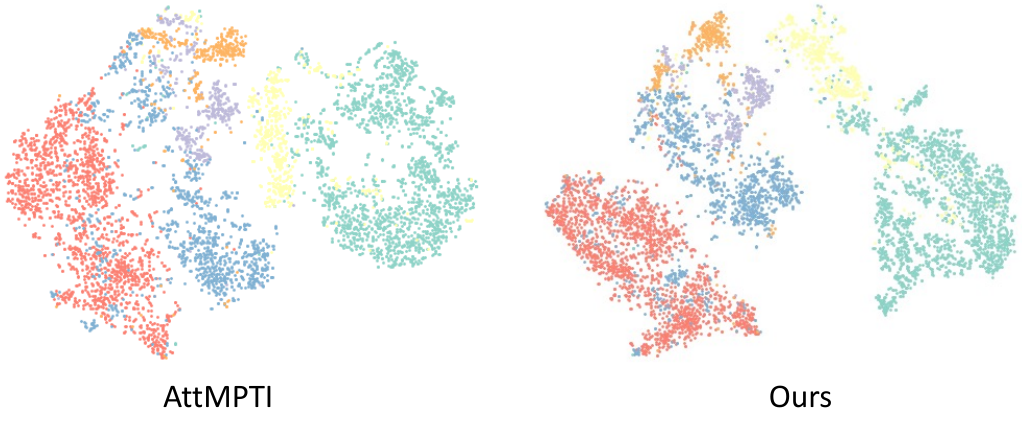}
\caption{t-SNE \cite{van2008visualizing} comparison on the \textbf{testing classes} of S3DIS.}
\label{tsne-s3dis}
\end{figure}

Fig.~\ref{tsne-s3dis} presents visualization of the feature distribution of testing classes on the S3DIS via t-SNE \cite{van2008visualizing}. Each color represents a different class. By learning with our proposed component-level clean noise separation, the intra-class is more compact and inter-class is more separable.  

\begin{figure}[h]
\centering
\includegraphics[scale=0.55]{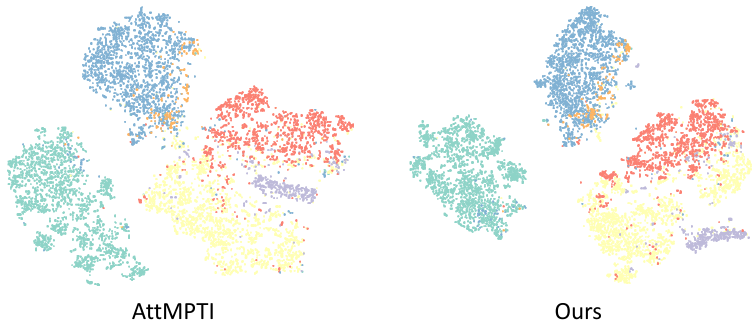}
\caption{t-SNE \cite{van2008visualizing} comparison on the \textbf{training classes} of S3DIS.}
\label{tsne-s3dis-train}
\end{figure}

Fig.~\ref{tsne-s3dis-train} presents visualization of the feature distribution of training classes on the S3DIS via t-SNE \cite{van2008visualizing}.
Different colors represent different classes. It is obvious that the class-wise feature distribution of our method is more distinctive and compact than AttMPTI. Together with Fig.~\ref{tsne-s3dis}, we can conclude that our feature representation learning is able to not only make feature embedding of seen classes discriminative but also generalize to unseen classes. 

\section{Experiment Results on ScanNet}
\paragraph{Effectiveness of CCNS and MDNS.}
We analyze the effectiveness of our proposed component-level clean noise suppression (CCNS) and multi-scale degree-based noise suppression (MDNS) on ScanNet in Tab.~\ref{scannet-cl+cd}. Both CCNS and MDNS are effective, and the combination of them achieves best overall performance. It is worth highlighting that the robustness of AttMPTI is improved by simply adding our feature representation learning, \ie CCNS. It verifies our claim that AttMPTI has the potential to be noise robust (by FPS based multi-prototype generation and label propagation), yet is subject to how discriminative the feature embedding is.

\begin{table}[h]
\centering
\setlength{\abovecaptionskip}{0.1cm}
\resizebox{0.5\linewidth}{!}{
\begin{tabular}{c|c|cc|cc}
\toprule
\multirow{2}{*}{model} & \multirow{2}{*}{0\%} & \multicolumn{2}{c|}{In-episode Noise} & \multicolumn{2}{c}{Out-episode Noise} \\ \cline{3-6} 
                       &                      & 20\%              & 40\%              & 40\%              & 60\%              \\ 
\midrule
AttMPTI                & 54.16              & 46.63             &31.57                  & 43.31        & 36.45        \\ 
AttMPTI+CCNS          & \textbf{54.79}                     &47.80             &32.92                 & 45.54                  & \textbf{38.57}   \\
\textbf{Ours}                   &53.50                      & \textbf{49.78}             &\textbf{38.70}                 &\textbf{47.90}        &38.42   \\
\bottomrule
\end{tabular}
}
\caption{Effectiveness of CCNS and MDNS on the ScanNet on 2-way 5-shot. `Ours' consists of both CCNS and MDNS.}
\label{scannet-cl+cd}
\vspace{-5mm}
\end{table}

\begin{figure}[t]
\centering
\includegraphics[scale=0.38]{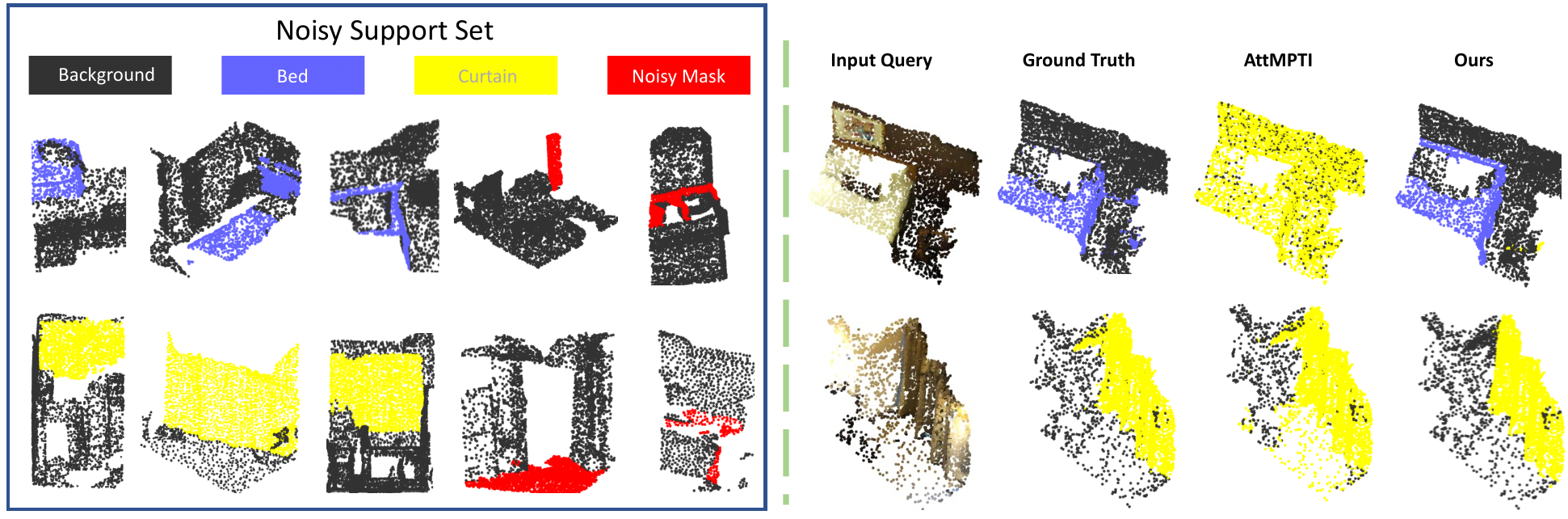}
\vspace{-3mm}
\caption{Qualitative comparison between AttMPTI and our method under a 2-way 5-shot point cloud segmentation with 40\% out-episode noise on ScanNet. Each row shows the segmentation results of a target class from the corresponding support set in the 2-way setting.}
\label{scannet-seg}
\end{figure}

\paragraph{Qualitative Results.}
Fig.~\ref{scannet-seg} presents the qualitative comparison between our method and AttMPTI under a 2-way 5-shot point cloud segmentation with 40\% out-episode noise on ScanNet \cite{dai2017scannet}. With the interference of the noisy shots, AttMPTI \cite{zhao2021few} either fails to segment the target semantic object (see the result in the first row) or wrongly segment some background points as the target class (see the result in the second row). In contrast, our method is able to give reliable segmentation results with respect to the target classes.

\section{Data split}
We follow the data split of \cite{zhao2021few}, and adopt the split 0 as the testing classes as shown in Tab.~\ref{data split}.

\begin{table}[h]
\centering
\resizebox{0.5\linewidth}{!}{
\begin{tabular}{c|c|c}
\toprule
Dataset & Meta-Testing Classes     & Meta-Training Classes                   \\ \midrule
\textbf{S3DIS}   & \begin{tabular}[c]{@{}c@{}}beam, board, bookcase, \\ ceiling, chair, column\end{tabular}                                   & \begin{tabular}[c]{@{}c@{}}door, floor, sofa, \\ table, wall, window\end{tabular}                                                          \\ \hline
\textbf{ScanNet} & \begin{tabular}[c]{@{}c@{}}bathtub, bed, bookshelf, \\ cabinet, chair, counter, \\ curtain, desk, door, floor\end{tabular} & \begin{tabular}[c]{@{}c@{}}otherfurniture, picture, refrigerator, \\ shower curtain, sink, sofa, \\ table, toilet, wall, window\end{tabular} \\ \bottomrule
\end{tabular}
}
\caption{Data split in S3DIS and ScanNet.}
\label{data split}
\end{table}

\section{Clean Ratio Comparison}
Tab.~\ref{clean ratio s3dis} lists the clean ratios
of the original support set (`Original') and the filtered support set produced by the MDNS (`Ours') during meta-testing. 
The clean ratio in each noise setting is given by first computing the percentage of the number of the clean shots in the corresponding set of one episode and then averaging the percentages in all episodes.  
As can be clearly seen from Tab.~\ref{clean ratio s3dis}, our method can significantly improve the clean ratio in all the noise setting.

\begin{table}[h]
\centering
\setlength{\abovecaptionskip}{0.1cm}
\resizebox{0.5\linewidth}{!}{
\begin{tabular}{p{1.5cm}<\centering|p{1.5cm}<\centering p{1.5cm}<\centering|p{1.5cm}<\centering p{1.5cm}<\centering}
\toprule
\multirow{2}{*}{Model} & \multicolumn{2}{c|}{In-episode Noise} & \multicolumn{2}{c}{Out-episode Noise} \\ \cline{2-5} 
                                    & 20\%              & 40\%              & 40\%              & 60\%              \\ \midrule
Original             & 0.8000            & 0.6000            & 0.6000            & 0.4000            \\
\textbf{Ours}                & \textbf{0.9749}           & \textbf{0.8211}            & \textbf{0.8476}            & \textbf{0.5602}            \\
\bottomrule
\end{tabular}
}
\caption{Comparison of the clean ratio of the 2-way 5-shot noisy support set in the meta-testing stage on the S3DIS.}
\label{clean ratio s3dis}
\end{table}

\section{Baseline Setups}
We compare our method with few-shot point cloud semantic segmentation (3DFSSeg) methods AttMPTI \cite{zhao2021few} and ProtoNet \cite{snell2017prototypical}, robust few-shot learning (R2DFSL) method Tra-NFS \cite{liang2022few} and robust point cloud semantic segmentation (R3DSeg) method PNAL \cite{ye2021learning}. All methods use the same feature extractor as AttMPTI for fair comparison. 

We follow the official code in AttMPTI to train ProtoNet and AttMPTI. For Tra-NFS, we adopt three-layer transformer encoder to generate robust prototype. We also randomly inject noise into the support set by sampling point clouds containing foreground objects from other classes during meta-training. For PNAL, we apply its robust training algorithm on each noisy support set and then test the performance on the corresponding query point cloud in each episode during meta-testing. We do not carry forward the knowledge from one episode to the next as suggested in \cite{mazumder2021rnnp}.

\bibliography{bmvc_review}
\end{document}